\documentclass[letterpaper, 10 pt, conference]{ieeeconf}  

\IEEEoverridecommandlockouts                              

\usepackage{amsmath} 

\usepackage{xcolor}
\usepackage{lipsum}
\usepackage{cuted}
\usepackage{capt-of}
\usepackage[utf8]{inputenc}
\usepackage{dirtytalk}
\usepackage[normalem]{ulem}
\usepackage{soul}
\usepackage{algorithm2e}
\usepackage{todonotes}
\usepackage{hyperref}
\usepackage{cite}

\usepackage{graphicx}

\usepackage{xspace}

\newcommand{\etal}{et al.\xspace}

\newcommand{\sect}{Sec.\xspace}
\newcommand{\fig}{Fig.\xspace}

\usepackage{textcomp}

\title{\LARGE \bf
REFORM: Recognizing F-formations for Social Robots
}

\author{
Hooman Hedayati$^{1}$,
Annika Muehlbradt$^{1}$,
Daniel J. Szafir$^{1}$,
Sean Andrist$^{2}$
\thanks{
$^{1}$ Department of Computer Science and ATLAS Institute, University of Colorado Boulder.
{\tt\small \{\href{mailto:hooman.hedayati@colorado.edu}{\color{blue}hooman.hedayati}, \href{mailto:annika.muehlbradt@colorado.edu}{\color{blue}annika.muehlbradt},\href{mailto:daniel.szafir@colorado.edu}{\color{blue}daniel.szafir}\}@colorado.edu}}
\thanks{$^{2}$Microsoft Research, Redmond 
{\tt\small \href{mailto:sandrist@microsoft.com}{\color{blue}sandrist@microsoft.com}}}
}

\begin{document}

\maketitle



\thispagestyle{empty}
\pagestyle{empty}


\begin{abstract}
Recognizing and understanding conversational groups, or \textit{F-formations}, is a critical task for situated agents designed to interact with humans. F-formations contain complex structures and dynamics, yet are used intuitively by people in everyday face-to-face conversations. Prior research exploring ways of identifying F-formations has largely relied on heuristic algorithms that may not capture the rich dynamic behaviors employed by humans. We introduce \textit{REFORM} (\textbf{RE}cognize F-\textbf{FOR}mations with \textbf{M}achine learning), a data-driven approach for detecting F-formations given human and agent positions and orientations. REFORM decomposes the scene into all possible pairs and then reconstructs F-formations with a voting-based scheme. We evaluated our approach across three datasets: the SALSA dataset, a newly collected human-only dataset, and a new set of acted human-robot scenarios, and found that REFORM yielded improved accuracy over a state-of-the-art F-formation detection algorithm. We also introduce \textit{symmetry} and \textit{tightness} as quantitative measures to characterize F-formations.

Supplementary  video: \href{https://youtu.be/Fp7ETdkKvdA}{\color{blue}https://youtu.be/Fp7ETdkKvdA}

Dataset available at: \href{https://github.com/cu-ironlab/Babble}{\color{blue}github.com/cu-ironlab/Babble}\\

\end{abstract}

\section{Introduction}
\label{sec::introduction}

Recent advances in AI, including developments in machine learning, natural language processing, computer vision, and dialogue systems, are fuelling the development of new socially interactive robots. These agents are rapidly increasing in popularity as researchers find new ways to integrate them into everyday life. For example, robots can assist humans in a variety of customer service tasks, such as welcoming, guiding, taking orders, and delivering items in shopping centers, hospitals, restaurants, hotels, and so on \cite{acosta2006design, datta2011pilot, kanda2009affective, osawa2017real, zalama2014sacarino, ahn2018robotic, scassellati2018teaching}.

Despite these promising advances, several challenges remain in developing robots that are able to socially interact with humans in a natural manner. 
Robots that are designed to interact with groups must take into account who is part of a conversation, who might be trying to join, and who might be leaving the group \cite{bohus2009dialog}. 


In this paper, we specifically focus on how robots might detect the spatial configurations of conversational groups. Prior research in the social sciences has explored several nuances regarding the structure and dynamics of social groups to understand how people organize and regulate group interactions. \cite{ciolek1980environment,kendon199213,kendon1990conducting,hedayati2020comparing}. Kendon operationalized the spatial configurations often formed in multi-party conversational groups as \textit{Facing Formations} or \textit{F-formations}: \say{An F-formation arises whenever two or more people sustain a spatial and orientational relationship in which the space between them is one to which they have equal, direct, and exclusive access} \cite{kendon1990conducting}. Since then, researchers have applied the concept of F-formations to other domains. For example, in human-computer interaction, F-formations have been used to inform how content can be shared across devices, while research in human-robot interaction has explored how F-formations can be used to guide robot behavior \cite{marquardt2012cross,vazquez2017towards}.

To date, detecting F-formations remains a challenging and open-ended problem. Most state-of-the-art algorithms are not robust because they rely on hard-coded parameters whose values are derived by experimentation and may not be generalizable. Existing algorithms use parameters such as \textit{stride}, the expected distance between an individual and the centroid of the F-formation. For example, the Graph-Cuts for F-formations (GCFF) algorithm \cite{setti2015f}, a state-of-the-art approach for detecting F-formations, uses \textit{stride} = 0.7 meters for the Cocktail Party dataset \cite{zen2010space}, but uses \textit{stride} = 0.5 meters for the CoffeeBreak dataset \cite{setti2015f}. 


We address these limitations and present three novel contributions. First, we introduce REFORM (\textbf{RE}cognize F-\textbf{FOR}mations with \textbf{M}achine learning), a new, data-driven method for reasoning about F-formations that does not rely on tuning parameters. REFORM consists of three phases: deconstruction, classification, and reconstruction. Unlike many data-driven methods, our approach does not require a large training dataset (in this work we show how REFORM's deconstruction phase can render 61,200 data points for training from a set of 400 annotated frames), which makes REFORM a viable approach for detecting F-formations.

Second, we provide to the research community a new dataset for exploring F-formations, called \textit{Babble}, in which participants played an improvisational word game in collocated groups. After training the REFORM algorithm on a single dataset, we provide the performance data of REFORM across three F-formation datasets: Babble, SALSA---an open existing dataset for multi-model interaction, and a \textit{HRI Proof-of-Concept Dataset} that consists of new data we collected from scripted interactions between a robot and researchers. We discuss these datasets in the Datasets section (\sect~\ref{sec::datasets}).

Third, we introduce \textit{Symmetry} and \textit{Tightness} as two new metrics for characterizing F-formations. Currently, F-formations are typically only described by the number of people in a group. However, group size alone may be insufficient for understanding important differences in F-formations. Conversational groups can be affected by physical objects (e.g., a table), limited space (e.g., a crowded bar), or context (e.g., presenter and audience), which can impact the shape and pattern of the resulting F-formation. Characterizing F-formations solely by the number of people may result in many F-formations appearing similar on the surface even when they differ in underlying structure and dynamics. Symmetry and Tightness provide new metrics for understanding F-formations and their variations. We discuss Symmetry and Tightness in detail in the F-Formation Characteristics section (\sect~\ref{sec::characterization}).


\section{Related Work}
\label{sec::relatedwork}

In recent years, researchers have developed several algorithms aimed at detecting F-formations, such as the GCFF algorithm \cite{setti2015f}, GRUPO \cite{vazquez2015parallel}, and GroupTogether \cite{marquardt2012cross}. Most of these algorithms use optimization techniques to identify the \textit{transactional segment}, the space in front of people where the interactions in the world take place, and the \textit{o-space}, a joint interaction space between people who are engaged in a group conversation \cite{ciolek1983proxemics}. Prior approaches commonly rely on using head orientations as a way to find \textit{transactional segments} under the assumption that human head orientation correlates with attentional focus \cite{setti2015f, cristani2011social}. However, head orientation data may be noisy and unreliable as people often move their heads while in conversational groups, for instance while looking at an active speaker, nodding to provide back-channel communication, or making gaze aversions to manage the conversational floor. 

Due to variations in human head movements while engaged in conversations, other research has explored body orientations as a potentially more reliable method for detecting F-formations. Vazquez et al. \cite{vazquez2015parallel} developed a method to estimate lower body orientation for F-formation detection. Marquardt et al. \cite{marquardt2012cross} developed another F-formation detection algorithm using data from a pair of Kinect depth cameras mounted on the ceiling. In this approach, the depth data is filtered and normalized using heuristic thresholds such that human head and body poses appear as two distinct ellipses. The F-formation is then classified based on the proximity and direction of these ellipses. Finally, Luber et al. \cite{luber2013multi} used people's direction of motion to detect transactional segments which provides useful data for moving groups, but is of limited value for static groups. 

Prior approaches to estimating the \textit{o-space} of an F-formation are based on Hough-voting strategies. For example, Cristani et al. developed Hough Voting for F-formations (HVFF) which first approximates the transactional segments using a Gaussian probability density function and then each function votes for an \textit{o-space} center to find the local maxima \cite{cristani2011social}. Setti et al. further improved this approach by using a multi-scale extension of Hough-voting \cite{setti2013multi}. Other methods for detecting the \textit{o-space} include Dominant Sets \cite{hung2011detecting} or Interacting Group Discovery \cite{tran2013social}.

While promising, many of these previous approaches for detecting F-formations are limited due to a reliance on threshold values determined heuristically using trial-and-error methods and tuned for specific datasets. 
For example, the GCFF algorithm \cite{setti2015f} uses a different threshold for \textit{stride} across various datasets 
to boost the accuracy of the algorithm. Moreover, using a single variable such as the transactional segment (which is a function of \textit{stride} and the position or orientation of a person) may result in low accuracy. The presence or absence of an F-formation depends on many factors including the distance between people, the velocity of people (e.g., standing still vs. walking), and the ease at which a person can look at others. Considering only one of these factors can cause algorithms to incorrectly detect F-formations (i.e., false positives or false negatives). For example, two people may be in close proximity to one another (distance) and looking at each other (head orientation) while in a conversational group or in passing; the latter does not constitute an F-formation.

As an alternative, our work uses a data-driven approach. We extend recent work examining machine learning models that can learn how to detect F-formations from data, rather than using hard-coded parameters. Mead et al. \cite{mead2013automated} demonstrated an early data-driven method for detecting F-formations of a fixed size. One limitation of this approach was that new models needed to be trained to detect F-formations of varying sizes. Gedik et al. \cite{gedik2018detecting} detected F-formations using ``Group-based meta-classifier learning using local neighborhood training'' (GAMUT), a data-driven approach that leverages information gathered from wearable devices attached to participants. Swofford \etal \cite{swofford2020improving} used data-driven approach with continuous likelihood for people interacting together to reason about the F-formations.
Below, we describe our approach with REFORM that can be used across different datasets, including new datasets, and only requires the positions and orientations of group members.

\section{REFORM}
\label{sec::reform}

REFORM (\textbf{RE}cognize F-\textbf{FOR}mations with \textbf{M}achine learning) is a data-driven approach 
based on an extension of our previous prototype for F-formation detection \cite{hedayati2019recognizing}. REFORM is first trained on a labeled dataset consisting of annotated frames, wherein each frame contains the positions and orientations of the people in the frame. Typically, such datasets are produced via manual human annotation because issues including camera perspective, uneven illumination, movement, and occlusions continue to pose challenges for automated approaches towards generating ground-truth labels. As manual annotation is a laborious process, one benefit of REFORM is that REFORM training requires a relatively small number of annotated frames due to our unique data deconstruction approach, described in detail below. After training, REFORM can detect F-formations (of any size) within any single frame by considering the position and orientation of individuals. REFORM outputs all possible F-formations in real time and can iteratively identify F-formations in any number of frames. REFORM consists of three steps: (1) Data Deconstruction, (2) Pairwise Classification, and (3) Reconstruction.

\subsection{Dataset Deconstruction}

Our first step is to deconstruct the training data to increase the number of data points in our training dataset, enabling REFORM to be trained on a relatively small number of annotated frames. This is an important step as obtaining large, labeled F-formation datasets remains challenging due to the laborious and inefficient nature of manual data annotation. To make the most of our annotated data, we deconstruct each frame of $n$ individuals and break these into pairs of two to create $n(n-1)/2$ pairwise data points. Each data point contains information about the head and body position and orientation of two individuals. Deconstructing our data into pairwise data points yields a greater variety of formations (both F-formations and non-F-formations) than looking at each frame as a whole.

\subsection{Pairwise Classification}
In the second step, REFORM identifies F-formations using feature-based classification. Specifically, we have selected two features to describe F-formations: \textit{Distance} and \textit{Effort Angle}. Distance is commonly used in F-formation detection and is defined as the Euclidean distance between two individuals. This is an intuitive measure because people commonly converse with others in close proximity. In addition to distance, we introduce a new metric of \textit{Effort Angle}. Effort Angle (\textit{EA}) defines how much body rotation would be required for two people to face each other with their bodies directly pointed at one another. $EA$ range is between 0--$2\pi$ where 0 indicates that two people are directly facing each other and $2\pi$ indicates that they are facing in opposite directions. $EA$ is commutative ($EA_{12} = EA_{21}$).

\textbf{\textit{Training:}} Although we explain the training phase here, we emphasize that training only happened once on a single dataset and that the trained classifiers are then used in the classification phase across several datasets. To train REFORM's classifiers, we used ~60\% of the SALSA dataset \cite{alameda2015salsa} (explained in detail in the \sect~\ref{sec::datasets}), roughly equal to 400 frames. Each of the 400 frames contained 18 people. We deconstructed the training set to extract all of the possible pairwise combinations from each frame ($18 \times 17 / 2 = 153$ pairs), resulting in 61,200 total pairwise datapoints. We encoded each datapoint with \textit{Distance}, \textit{Effort Angle}, and a classification label (in an F-formation or not). Next, we applied three classification methods to the pairwise data: Weighted KNN, Bagged Trees, and Logistic Regression. 

We selected these classifiers to sample performance across several types of models as they each belong to a different category of classifier. Weighted KNN is a K-nearest Neighbor classifier, Bagged Trees is an ensemble method, and Logistic Regression is type of Regression classifier. In REFORM, classifiers are exchangeable, allowing us to experiment with different classifiers and select the best performing classification method.

\textbf{\textit{Classification:}} To classify new data, we calculate the \textit{Distance} and \textit{Effort Angle} features for each pairwise data point. Then, we feed each data point to one of three classifiers (Weighted KNN, Bagged Trees, and Logistic Regression). The resulting predictions are used to create a \textit{Relation Matrix ($M_R$)}. For $n$ people in a frame, $M_R$ is a $n\times n$ symmetric matrix and for each pair of two people $P_i$ and $P_j$, the elements $M_{R_{ij}}$ and $M_{R_{ji}}$ are equal to the predicted label and the diagonal entries are all $1$.

\subsection{Reconstruction}
The final step is to reconstruct the pairwise data into sets of F-formations. Because classifiers are not without error, we expect to see some inconsistencies across the pairwise predictions. For example, in a frame with three individuals $P_1$, $P_2$, and $P_3$, pairwise classifications might indicate that $P_1$ and $P_2$, and $P_2$ and $P_3$ are in F-formations but $P_1$ and $P_3$ are not in an F-formation. 

To resolve this issue, we implemented \textit{Greedy Reconstruction} (Algorithm \ref{reconstruction}), which aggregates pairwise F-formations only when the majority of the predictions indicate that a larger F-formation exists across the pairs. Let the row $i$ in $M_R$ indicate predictions about whether a person $P_i$ is in an F-formation with respect to others (i.e., all of the other people in the scene who could possibly be in an F-formation with $P_i$). We refer to the set of classifier predictions (beliefs) in row $i$ as $B_i$. The Greedy Reconstruction algorithm finds $P_i$ and $P_j$ that have the maximum number of elements in their belief intersection ($B_i$ $\cap$ $B_j$). Then, $B_i$ $\cup$ $B_j$ constitutes an F-formation and $B_i$ and $B_j$ are deleted from $M_R$. This process repeats until there is no $B$ left in $M_R$. For example, take $B_1$={$P_1$,$P_2$,$P_3$}, $B_2$ = {$P_1$,$P_3$,$P_4$}, and $B_3$={$P_1$,$P_2$,$P_3$}. It is more likely that $B_2$ has been incorrectly identified as an F-formation compared to $B_1$ and $B_3$. In other words, it is more likely that the classifier made one mistake rather than two.

\begin{figure*}[h]
  \includegraphics[width=\textwidth]{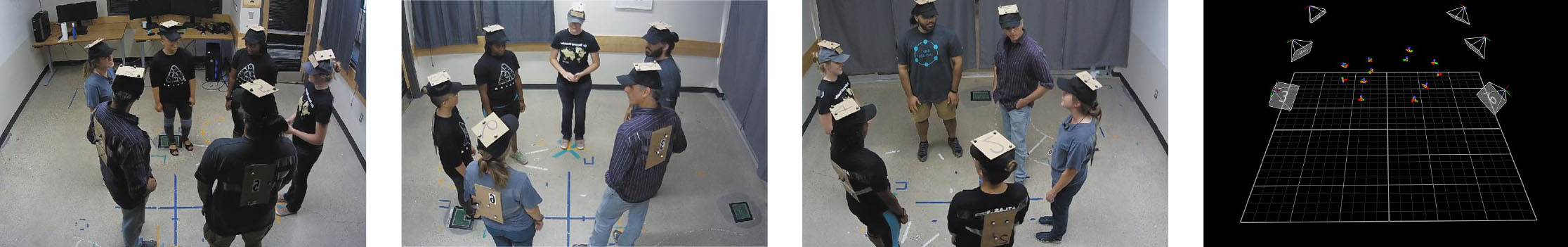}

  \caption{Examples of the reflective markers used to track participants, different F-formations captured by the RGB cameras, and head and body positions and orientations captured by the Vicon cameras used to construct the Babble dataset.}
  \label{fig:babble}
\end{figure*}

\vspace{10pt}
\begin{algorithm}
GreedyReconstruction ($M_R$):\\
\vspace{5pt}
$\Psi = \{\}$ \hspace{5pt} // \small{\textit{The F-formation set}}\\
\While{Size($M_R$) $\geq$ 2}{
    \For{$\forall P_i  \forall P_j$ in $M_R$}{
        {$Agreement_{ij} \gets Size ( B_i \cap B_j)$}\\
        \If{$P_i \not\in (B_i \cap B_j) \lor P_j  \not\in (B_i \cap B_j)$}{
        $Agreement_{ij} \gets Agreement_{ij} \cup P_i \cup P_j$\\
        }
        
        $\Psi \gets Max(Agreement)$\\
        }
  ${M_R \gets M_R - (P_{i_{max}} \cup P_{j_{max}})}$
  }
return $\Psi$
\vspace{5pt}
\caption{Reconstruction Algorithm}\label{reconstruction}
\end{algorithm}

\section{Datasets}
\label{sec::datasets}
In this section, we describe the datasets we used to train and evaluate REFORM. We explain why we chose these datasets and describe their properties (e.g., number of frames, number of F-formations, etc.). We also describe the annotation process 
for the new datasets that we collected, namely the \textit{Babble} and the \textit{HRI Proof-of-Concept} datasets.

\subsection{SALSA Dataset}
SALSA \cite{alameda2015salsa} is an existing dataset consisting of a 60-minutes recording of social interactions between 18 individuals. In addition to the recording, SALSA contains position, pose, and F-formation annotations for every 3 seconds of data using a dedicated multi-view scene annotation tool to annotate the position, head orientation, and body orientation of each individual. 

Among all publicly available F-formation datasets, such as the Cocktail Party dataset \cite{zen2010space}, CoffeeBreak dataset \cite{cristani2011social}, MatchNMingle dataset \cite{zen2014unsupervised}, etc., we chose the SALSA dataset because it has a relatively large number of annotated frames and a large number of people in each frame. It includes up to 18 people per frame, contains frames with multiple F-formations, and frames with large conversational groups (with F-formations of up to 6 people). Among the annotated frames, we randomly divided the dataset into a training set (\texttildelow 60\%, corresponding to roughly 400 frames) and a testing set (\texttildelow 40\%, 245 frames). The training set was the only training data REFORM received, including when tested on other datasets.

One limitation of the SALSA dataset (and human annotated datasets in general) is that there are some inconsistencies and errors in the annotations (the SALSA data in particular has inconsistencies in the position and orientation annotations). To detect F-formations, it is important to have accurate values for the positions and orientations of people in all annotated frames. To address this issue and evaluate the generalizability of our approach, we also collected the \textit{Babble} dataset (described below).


\subsection{Babble Dataset}

One challenge of using a data-driven approach is that the trained model may correspond too closely to the limited set of data points on which it was trained on (i.e., ``overfitting the data'') and therefore performs well on the training set but poorly on new datasets. To evaluate the generalizability of our approach and provide a dataset with precise feature measurements, we collected a new dataset that we term the \textit{Babble} dataset. The Babble dataset consists of a 35-minute recording of conversational interactions between 7 individuals with precisely recorded head and body positions and orientations via a motion-tracking system and labeled F-formations from two annotators (a total of 3481 frames). 

While the Babble dataset has a similar duration of annotated data to SALSA, the number of participants is fewer (7 for Babble vs 18 for SALSA). However, Babble has a larger range of F-formations (groups of 2--7 people) than SALSA (2--6). In addition, the Babble dataset annotations take place at a higher frequency (SALSA annotations occur every 3 seconds vs every 0.5s in Babble) and the Babble data includes accurate head and body orientation annotations (recorded to within 1mm). This also sets Babble apart from the Cocktail Party \cite{zen2010space} and CoffeeBreak datasets \cite{cristani2011social}, which also contain inconsistencies in the annotations and less accurate head and body orientations. To our knowledge, Babble is the first dataset to provide absolute head and body orientation and labeled F-formations for all individuals in a frame. This dataset is publicly available on our \href{https://github.com/cu-ironlab/Babble}{\color{blue}git repository}\footnote{ \href{https://github.com/cu-ironlab/Babble}{\color{blue}github.com/cu-ironlab/Babble}\\}.

To collect the Babble dataset, we recruited 7 students (3 female, 3 male, 1 non-binary) from our University campus by word of mouth under a protocol approved by our local IRB. We conducted a 1-hour experiment in our lab in which participants played a social game called the Improv Word Game. First, we outfitted participants with reflective  markers and asked each participant to briefly stand inside of our $5m \times 5m \times 3m$ game area in order to calibrate our motion tracking cameras. Then, we introduced participants to the Word Improv Game and played one practice round to ensure participants understood how to play. Finally, participants played the game until there was a winner (\texttildelow 35 minutes).

\begin{figure*}[h]
  \includegraphics[width=\textwidth]{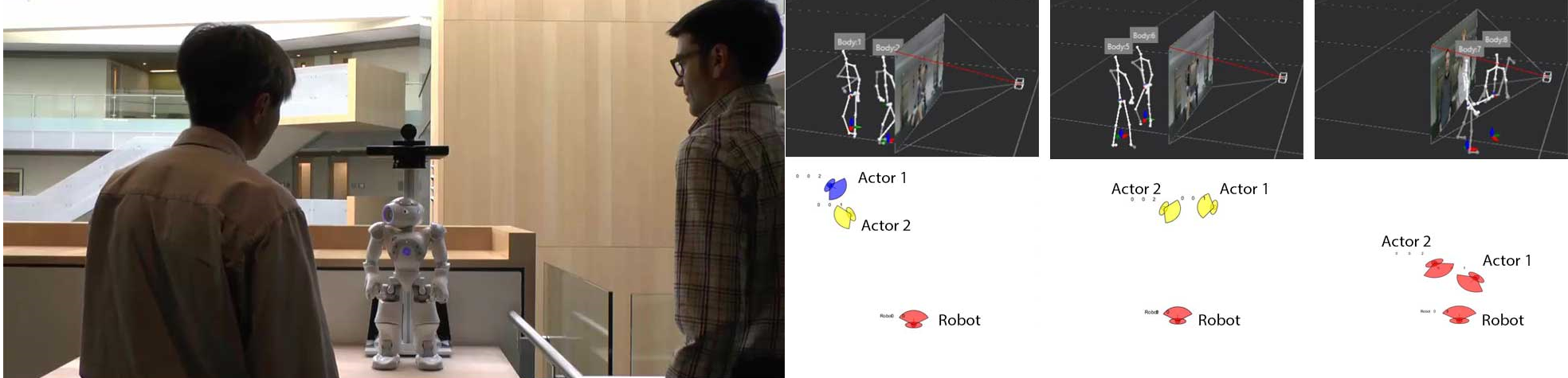}
  \caption{The setup used for the HRI Proof-of-Concept Dataset wherein two researchers are in an F-formation with a robot (left); examples of the 3 different F-formations: no interaction (no F-formation), a group conversation between two humans (F-formation of size 2), and a group conversation between two humans and a robot (F-formation of size 3) (right).}
  \label{fig:HRI}
\end{figure*}

\textbf{\textit{Word Improv Game:}} This activity had participants play a game with a deck of cards, where each card had a random word on it, a moderator, and a moderator. The game comprises a deck of cards with a random word on each card, a moderator, and two or more players.  The moderator and players stand in a circle. The moderator picks the first card in the deck, reads the word aloud, and chooses a player for the turn. The chosen player must talk about the given word for roughly 30 seconds without pausing. Players can talk about anything that is associated with the word. When 30 seconds are up, the players ``shuffle'' by randomly switching places with one another so that the arrangement of players in the circle differs from the previous round. Then, the moderator chooses a new word and a player for the next turn. If a player is unable to talk about a word for 30 seconds, or strays off-subject, the player loses and is out of the game. When a player is out, they must leave the circle and the remaining players shuffle. The winner is the player that is left after all other players are out.

To track participants' position and orientation throughout the game, we outfitted each participant with 6 IR reflective markers. Three markers were attached to participants' backs to track the position and orientation of their torsos and three markers were attached to a baseball cap to track the position and orientation of participants' heads. To track the markers, we used 6 Vicon motion capture cameras that captured participants' head and body orientations at 100 frames per second with a precision of 1 mm. Additionally, we mounted 3 RGB cameras on the walls of the room to capture images of the participants at 30 frames per second. The setup can be seen in \fig~\ref{fig:babble}.

In total, we recorded 3481 frames, with 3 RGB images (used for F-formation annotations) and head and body orientation data via motion capture for each frame. To generate the Babble dataset, we combined static RGB images taken at 0.5 second intervals with the tracking information (position and orientation) captured by the Vicon cameras. Then, two members of the research team manually annotated each of the frames. The annotators separately categorized the frames by specifying the F-formation memberships (e.g., \{2,3,5\}, \{1,4\} for two F-formations involving participants 2, 3, and 5 in the first, and participants 1 and 4 in the second). If no participants were in an F-formation, the annotation is empty. Frames in which players were shuffling typically did not contain any F-formations as players were walking around randomly. Transitions, such as a player leaving the game, were typically characterized by F-formations because the remaining players either preserved the existing formation or adjusted their positions to form a new F-formation. We compared the annotations for F-formation memberships across the two annotators; the inter-rater reliability (Cohen's kappa) was $\kappa$ = 0.82.

\subsection{HRI Proof-of-Concept Dataset}

In addition to the new Babble dataset, we also collected data for another dataset due to our underlying motivation of improving F-formation detection for social robots. To evaluate REFORM's performance in human-robot interactions, we collected a small dataset that we will refer to as the \textit{HRI Proof-of-Concept Dataset}. This dataset consists of a 5-minute recording of scripted interactions between the ``Directions Robot'' \cite{bohus2014} and a member of our research team and a colleague. It also includes frame-by-frame annotations for F-formations along with the humans' positions and orientations automatically provided by a Kinect depth sensor (a total of 3091 frames). While this is a small dataset, it includes data about F-formations consisting of both humans and a robot (unlike SALSA or Babble). 

\textbf{\textit{Directions Robot:}} The Directions Robot is a small humanoid that functions as a directional guide by providing walking instructions to buildings, offices, and other public areas. It supports naturalistic conversation and can respond to questions (e.g., ``where is John's office?'') using natural language and gestural output \cite{bohus2014}. We chose the Directions Robot because it supports conversational interactions between one or more individuals and because it is already in use at Microsoft Research to guide visitors.

To collect the \textit{HRI Proof-of-Concept Dataset}, we enacted 3 different scenarios as shown in Figure \ref{fig:HRI}: (1) a group conversation with two humans and the robot, (2) a group conversation between two humans (robot excluded), and (3) no interactions between humans and robot. In the first scenario, the researchers and robot were engaged in conversation and the researchers asked the robot questions about a room in the building. The researchers and robot formed an F-formation of size 3. In the second scenario, the two researchers stood in front of the robot and engaged in a short conversation with each other. Both the robot and the researchers are in the scene but only the two researchers formed an F-formation. In the third scenario, researchers walked past each other multiple times in front of the robot. The interaction did not constitute any F-formation.

To track the researchers, we attached a Kinect and a wide-angle RGB camera atop the Directions Robot. We then used skeleton tracking to calculate the researchers' positions and orientations with respect to the robot. 

We recorded a total of 3091 frames during this session. To generate the dataset, we randomly selected 100 frames and two members of the research team separately annotated each with the F-formation membership. 

\section{Evaluation}
\label{sec::evaluation}

\begin{table*}[htbp]
\begin{center}
\caption{REFORM's performance on training and testing data (\texttildelow 60\% and \texttildelow 40\% of the SALSA dataset, respectively) compared to a majority baseline classification and the GCFF algorithm.}
\begin{tabular}{|c|c|c|c|c|c|c|c|}
\hline
& \textbf{Pairwise Accuracy}& \textbf{Precision Train} & \textbf{Precision Test} & \textbf{Recall Train} & \textbf{Recall Test} & \textbf{F1 Train} & \textbf{F1 Test} \\
\hline
\textbf{Majority Baseline}& 85.3 & 100&100 & 0&0 & 0&0 \\
\hline
\textbf{Graph-Cuts}&N/A & 66.2&63.8 &64.2&64.2 &65.2&64.2 \\
\hline
\textbf{Weighted KNN}&92.1& 86.5&78.1 &99.9&82.7 &92.7&80.3 \\
\hline
\textbf{Bagged Tree} & \textbf{93.3} & \textbf{86.3} & \textbf{78.3} &\textbf{99.4}& \textbf{84.3} &\textbf{92.4} &\textbf{81.2} \\
\hline
\textbf{Logistic Regression} & 92.2  & 73.9&71.3 & 78.9&78.6 & 76.3&74.8 \\
\hline
\end{tabular}
\label{result-salsa}
\end{center}
\end{table*}

We evaluated REFORM on the three datasets described above (SALSA, Babble, and HRI Proof-of-Concept Dataset). For all of the results below, we emphasize that the REFORM classifiers were only trained once (as discussed in the Training section) using only the training set derived from 60\% of the SALSA data.

First, we evaluated REFORM's performance on the SALSA dataset across our three classifiers (Weighted KNN, Bagged Tree, and Logistic Regression) and compared our approach to the existing state-of-the-art GCFF algorithm \cite{setti2015f}. We tuned the parameters ($\textit{MDL} = 30000$ and $\textit{stride} = 0.7$) as specified by the GCFF open-source code to ensure a fair comparison. All three classifiers outperformed the GCFF algorithm by as much as 20\% in precision, recall, and F1 score. While further parameter tuning could potentially improve GCFF's performance, it is worth noting that REFORM achieved high performance without the need to tune any parameters. The Bagged Tree classifier had the best overall performance, however, Logistic Regression exhibited the least degree of overfitting. The precision, recall, and F1 scores for $T=2/3$ (i.e., predictions with a 2/3 match to ground truth are considered correct as described by Setti et al. \cite{setti2015f}) are shown in Table \ref{result-salsa}.   

Our second evaluation compared REFORM to the GCFF algorithm on our novel Babble dataset. Again, we emphasize that REFORM was trained only on the SALSA dataset and then tested on the Babble dataset (i.e., the classifiers were not re-trained on the new data). As before, we set GCFF's parameters to the specified values ($\textit{MDL} = 30000$ and $\textit{stride} = 0.7$). Again, all three classifiers outperformed the GCFF algorithm in precision, recall, and F1 score. The Bagged Tree classifier had the highest scores for all three measures and performed comparatively well on this dataset relative to the SALSA dataset (e.g., the classifier's Precision score was only 2.7\% lower than on the SALSA dataset). Precision, Recall, and F1 measures are summarized in Table \ref{results-resistance}. The results of the Bagged Tree classifier provide compelling evidence that REFORM is generalizable to a greater degree than previous approaches.

\begin{table}[b]
\begin{center}
\caption{REFORM's performance on the Babble dataset compared to the GCFF algorithm.}
\label{results-resistance}
\begin{tabular}{|c|c|c|c|}
\hline
& \textbf{Precision} & \textbf{Recall}& \textbf{F1} \\
\hline
\textbf{Graph-Cuts} & 65.0 & 75.0 &69.6 \\
\hline
\textbf{Weighted KNN} & 74.3 & 87.5 & 80.4 \\
\hline
\textbf{Bagged Tree}& \textbf{75.6} & \textbf{90} & \textbf{82.1} \\
\hline
\textbf{Logistic Regression} & 65.6 & 80.0 & 72.1 \\
\hline
\end{tabular}
\end{center}
\end{table}

Last, we tested our classifiers on our \textit{HRI Proof-of-Concept Dataset} and compared their performance to the GCFF algorithm. All three classifiers performed equally well on Precision (75.7), Recall (100), and F1 score (86.2), and once again outperformed the GCFF algorithm (Precision: 51.5, Recall: 63.6, F1: 56.9). While these results are promising, we note that this dataset is small and represents acted out, rather than purely natural human-robot interactions. It also only contains two humans and one robot, resulting in a handful of different F-formations to be classified. We speculate that if we increase the number of people who interact with the robot in our dataset and collect data using natural interactions, the accuracy of these classifiers will change.


\section{F-formation Characterization}
\label{sec::characterization}

\begin{figure*}
 \centering
   \includegraphics[width=\textwidth]{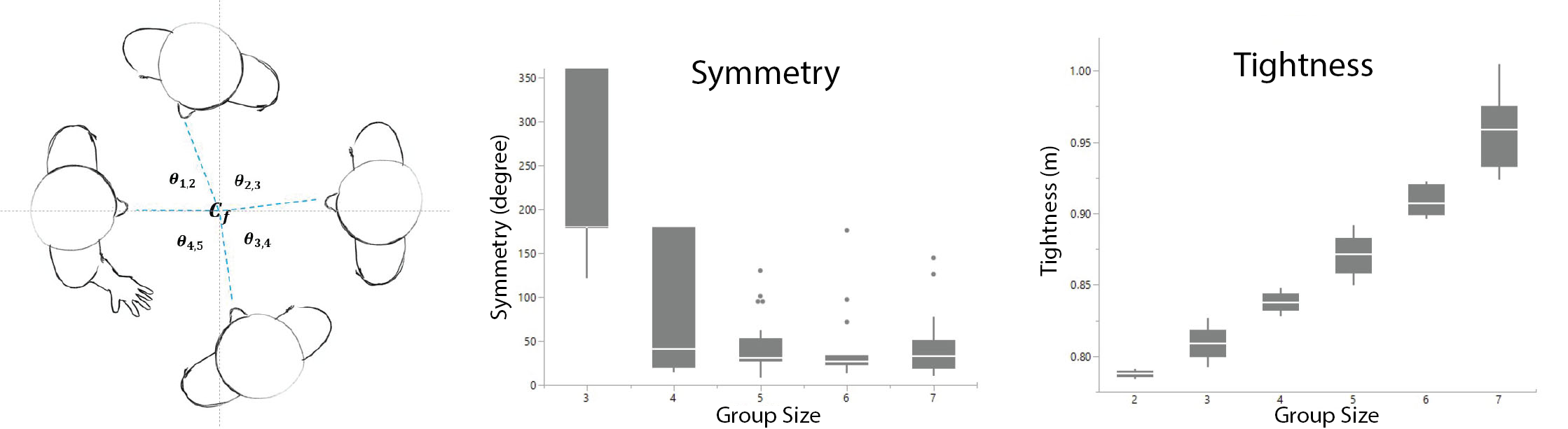}
   \caption{Left: \textit{Symmetry} measures the angles between individuals in a F-formation; Center: We found \textit{Symmetry} to be fairly consistent across F-formations size except for groups of 3; Right: We found \textit{Tightness} varied across F-formation size.}
   \label{fig:result}
 \end{figure*}
 
While we believe these initial results show that REFORM holds promise in a providing a new, data-driven method for generalizable F-formation detection, we are also interested in understanding how novel features may be useful in detecting and characterizing F-formations. F-formations are typically described by the number of people in a group. The size of a group, however, may be insufficient for detecting important differences in F-formations. For example, F-formations can be constrained by physical objects (e.g., a table), limited by available space (e.g., a crowded bar), or altered by context (e.g., presenter and audience), all of which can impact the shape and pattern of the formation. As described above, our REFORM method currently utilizes \textit{Distance} and \textit{Effort Angle} as features for classifying F-formations. In this section, we discuss two new ways to characterize F-formations that arose from our analysis of REFORM's performance that may help future work better distinguish between F-formations of different structures and dynamics.

In evaluating REFORM, we observed two interesting patterns in the formation of conversational groups with respect to F-formation size, \textit{Distance}, and \textit{Effort Angle}. First, we observed that F-formations that are unconstrained tend to be symmetric and circular in shape, wherein each person has a similar \textit{Distance} and \textit{Effort Angle} compared to their neighbor. We call this \textit{Symmetry}, a measure to describe the degree to which the angles between people in an F-formation are congruent. Second, we found that the average distance from the center of an F-formation to the individuals in the group appears to grow relative to its size (the total number of people in the F-formation). We call this \textit{Tightness}, the average Euclidean distance between group members and the center of the conversational group.

\textbf{Symmetry:} To describe Symmetry, we denote the angle between adjacent people $i$ and $j$ in a F-formation as $\theta_{i,j}$. In a perfectly symmetric F-formation of size N, all of the angles in $\theta_{i,j}$ would be equal, with $\theta_{i,j} = 360/N$. We refer to perfect Symmetry as $\theta_{Perfect}$. The Symmetry of F-formation $S_f$ is defined as the accumulated errors between pairs of adjacent people with $\theta_{Perfect}$. For example, if there are four individuals in an F-formation, there would be four adjacent angles $\theta_{1,2}$, $\theta_{2,3}$, $\theta_{3,4}$, and $\theta_{4,1}$, and $\theta_{Perfect} = 90$. The formal definition of Symmetry is as follows:
\begin{equation}
T_f = \sum_{i=1}^{N} {|\theta_{Perfect}-\theta_{i,j}|}
\end{equation}

We manually inspected 100 frames in the Babble dataset and found no differences in Symmetry for F-formations of size 4, 5, 6, and 7. However, we did find significant differences in Symmetry when comparing F-formations of size 3 to larger F-formations as determined by a one-way ANOVA using formation size as a fixed effect, $F(4,94) = 47.67$ , $p < .0001$. A post-hoc test using Tukey's HSD revealed that the degree of Symmetry was significantly lower for $f_3$ (M = 234) compared to $f_4$ (M = 87), $f_5$ (M = 44), $f_6$ (M = 42), and $f_7$ (M = 44), with $p < 0.001$. 
We speculate that this effect may in part be due to the gameplay of the Word Improv Game, where an F-formation of size 3 includes two opponents and one moderator. In our study, the two remaining opponents often faced each other (presumably due to the competitive nature of the task), which could lead them to exclude the moderator from the F-formation. Symmetry is not defined for a vis-a-vis arrangement as people are positioned in a straight line and maintain perfect symmetry, $S_{(f_2)}=0$. While there are some discrepancies in our results, we found that most of our unconstrained conversational groups (F-formations with more than three individuals) had a tendency to form F-formations with a high degree of Symmetry.

We observed similar patterns in Symmetry in the SALSA dataset. We inspected 500 frames and conducted a one-way ANOVA using formation size as a fixed effect, $F(4,1639) = 171.2$ , $p < .0001$. We found no differences in Symmetry for F-formations of size 4, 5, 6, and 7, but found significant differences when comparing size 3 to larger F-formations. A post-hoc test using Tukey's HSD showed that the degree of Symmetry was significantly lower for $f_3$ (M = 50) compared to $f_4$ (M = 36), $f_5$ (M = 32), $f_6$ (M = 32), and $f_7$ (M = 26), with $p < 0.001$. 

\textbf{Tightness:} The Tightness of an F-formation ($T_f$) is the Euclidean distance between the group members and the center of the conversational group. Tightness is similar to Setti's F-formation radius $R_k$  \cite{setti2013multi}, but differs in that $T_f$ does not use any predefined heuristic values (such as $s = 95cm$, where $s$ refers to the personal range that humans maintain between themselves). 

As with Symmetry, we explored Tightness by manually inspecting 100 frames in the Babble and SALSA datasets. We conducted a one-way ANOVA using the size of an F-formation as a fixed effect and found that F-formation size had a significant effect on Tightness, $F(5,94) = 258$, $p < .0001$. For the Babble dataset, a post-hoc test using Tukey's HSD revealed that all F-formations had significantly different Tightness: $f_2$ (M = .78), $f_3$ (M = .80), $f_4$ (M = .83), $f_5$ (M = .87), $f_6$ (M = .90), $f_7$ (M = .95), all $p < .0001$, except for F-formation size 2 and 3 for which $p = 0.0173$. 
For the SALSA dataset, all F-formations had significantly different Tightness except for F-formation size 5 and 6: $f_2$ (M =0.44), $f_3$ (M = 0.54), $f_4$ (M = 0.61), $f_7$ (M = 0.93), all $p < .0001$, and $f_5$ (M = 0.66), $f_6$ (M = 0.67). Our results show that as the size of an F-formation increases, Tightness decreases (i.e., people move further away from the F-formation center). 


We recognize that there are factors other than size that may influence the Symmetry and Tightness of an F-formation. In fact, these measures may help reveal important contextual differences in F-formations, such as differences in conversational groups with members of varying heights (e.g., adults and children), differences in the level of intimacy of the group members (e.g., close friends, colleagues), and level of noise in the environment (e.g., people tend to stand closer together in loud places). We intend to explore these factors in future research.





\section{Conclusion}
\label{sec::conclusion}

F-formation detection is important for the advancement of social robots. To support natural human-robot interaction, social robots need to be able to detect F-formations and recognize the differences in the shape and pattern of F-formations. This work contributes REFORM, a new data-driven approach for detecting F-formations that outperforms the state-of-the-art GCFF algorithm on existing and new datasets. Our approach appears to be generalizable and holds promise for detecting F-formations in the wild. We also present a new dataset for future research into social interactions that contains precise human orientation and positional information with annotated F-formations. Finally, we describe two new metrics for characterizing F-formations of different shapes and patterns: \textit{Symmetry} and \textit{Tightness}, which may help reveal important contextual differences across F-formations. 
Robots that can recognize these contextual differences will be able to enact more sophisticated interaction policies for smoother and more enjoyable user interaction.

\section{Acknowledgements}
This work was supported by an Early Career Faculty grant from NASA’s Space Technology Research Grants Program under award NNX16AR58G. We thank Benjamin Lee for his help in our research.


\bibliographystyle{IEEEtran}
\bibliography{refs}

\begin{thebibliography}{10}
\providecommand{\url}[1]{#1}
\csname url@samestyle\endcsname
\providecommand{\newblock}{\relax}
\providecommand{\bibinfo}[2]{#2}
\providecommand{\BIBentrySTDinterwordspacing}{\spaceskip=0pt\relax}
\providecommand{\BIBentryALTinterwordstretchfactor}{4}
\providecommand{\BIBentryALTinterwordspacing}{\spaceskip=\fontdimen2\font plus
\BIBentryALTinterwordstretchfactor\fontdimen3\font minus
  \fontdimen4\font\relax}
\providecommand{\BIBforeignlanguage}[2]{{%
\expandafter\ifx\csname l@#1\endcsname\relax
\typeout{** WARNING: IEEEtran.bst: No hyphenation pattern has been}%
\typeout{** loaded for the language `#1'. Using the pattern for}%
\typeout{** the default language instead.}%
\else
\language=\csname l@#1\endcsname
\fi
#2}}
\providecommand{\BIBdecl}{\relax}
\BIBdecl

\bibitem{acosta2006design}
L.~Acosta, E.~Gonz{\'a}lez, J.~N. Rodr{\'\i}guez, A.~F. Hamilton \emph{et~al.},
  ``Design and implementation of a service robot for a restaurant,''
  \emph{International Journal of Robotics \& Automation}, vol.~21, no.~4, p.
  273, 2006.

\bibitem{datta2011pilot}
C.~Datta, A.~Kapuria, and R.~Vijay, ``A pilot study to understand requirements
  of a shopping mall robot,'' in \emph{Proceedings of the 6th international
  conference on Human-robot interaction}.\hskip 1em plus 0.5em minus
  0.4em\relax ACM, 2011, pp. 127--128.

\bibitem{kanda2009affective}
T.~Kanda, M.~Shiomi, Z.~Miyashita, H.~Ishiguro, and N.~Hagita, ``An affective
  guide robot in a shopping mall,'' in \emph{Proceedings of the 4th ACM/IEEE
  international conference on Human robot interaction}.\hskip 1em plus 0.5em
  minus 0.4em\relax ACM, 2009, pp. 173--180.

\bibitem{osawa2017real}
H.~Osawa, A.~Ema, H.~Hattori, N.~Akiya, N.~Kanzaki, A.~Kubo, T.~Koyama, and
  R.~Ichise, ``What is real risk and benefit on work with robots?: From the
  analysis of a robot hotel,'' in \emph{Proceedings of the Companion of the
  2017 ACM/IEEE International Conference on Human-Robot Interaction}.\hskip 1em
  plus 0.5em minus 0.4em\relax ACM, 2017, pp. 241--242.

\bibitem{zalama2014sacarino}
E.~Zalama, J.~G. Garc{\'\i}a-Bermejo, S.~Marcos, S.~Dom{\'\i}nguez, R.~Feliz,
  R.~Pinillos, and J.~L{\'\o}pez, ``Sacarino, a service robot in a hotel
  environment,'' in \emph{ROBOT2013: First Iberian Robotics Conference}.\hskip
  1em plus 0.5em minus 0.4em\relax Springer, 2014, pp. 3--14.

\bibitem{ahn2018robotic}
H.~S. Ahn, S.~Zhang, M.~H. Lee, J.~Y. Lim, and B.~A. MacDonald, ``Robotic
  healthcare service system to serve multiple patients with multiple robots,''
  in \emph{International Conference on Social Robotics}.\hskip 1em plus 0.5em
  minus 0.4em\relax Springer, 2018, pp. 493--502.

\bibitem{scassellati2018teaching}
B.~Scassellati, J.~Brawer, K.~Tsui, S.~Nasihati~Gilani, M.~Malzkuhn, B.~Manini,
  A.~Stone, G.~Kartheiser, A.~Merla, A.~Shapiro \emph{et~al.}, ``Teaching
  language to deaf infants with a robot and a virtual human,'' in
  \emph{Proceedings of the 2018 CHI Conference on Human Factors in Computing
  Systems}, 2018, pp. 1--13.

\bibitem{bohus2009dialog}
D.~Bohus and E.~Horvitz, ``Dialog in the open world: platform and
  applications,'' in \emph{Proceedings of the 2009 international conference on
  Multimodal interfaces}.\hskip 1em plus 0.5em minus 0.4em\relax ACM, 2009, pp.
  31--38.

\bibitem{ciolek1980environment}
T.~M. Ciolek and A.~Kendon, ``Environment and the spatial arrangement of
  conversational encounters,'' \emph{Sociological Inquiry}, vol.~50, no. 3-4,
  pp. 237--271, 1980.

\bibitem{kendon199213}
A.~Kendon, ``13 the negotiation of context in face-to-face interaction,''
  \emph{Rethinking context: Language as an interactive phenomenon}, no.~11, p.
  323, 1992.

\bibitem{kendon1990conducting}
------, \emph{Conducting interaction: Patterns of behavior in focused
  encounters}.\hskip 1em plus 0.5em minus 0.4em\relax CUP Archive, 1990,
  vol.~7.

\bibitem{hedayati2020comparing}
H.~Hedayati, D.~Szafir, and J.~Kennedy, ``Comparing f-formations between humans
  and on-screen agents,'' in \emph{Extended Abstracts of the 2020 CHI
  Conference on Human Factors in Computing Systems}, 2020, pp. 1--9.

\bibitem{marquardt2012cross}
N.~Marquardt, K.~Hinckley, and S.~Greenberg, ``Cross-device interaction via
  micro-mobility and f-formations,'' in \emph{Proceedings of the 25th annual
  ACM symposium on User interface software and technology}.\hskip 1em plus
  0.5em minus 0.4em\relax ACM, 2012, pp. 13--22.

\bibitem{vazquez2017towards}
M.~V{\'a}zquez, E.~J. Carter, B.~McDorman, J.~Forlizzi, A.~Steinfeld, and S.~E.
  Hudson, ``Towards robot autonomy in group conversations: Understanding the
  effects of body orientation and gaze,'' in \emph{Proceedings of the 2017
  ACM/IEEE International Conference on Human-Robot Interaction}.\hskip 1em plus
  0.5em minus 0.4em\relax ACM, 2017, pp. 42--52.

\bibitem{setti2015f}
F.~Setti, C.~Russell, C.~Bassetti, and M.~Cristani, ``F-formation detection:
  Individuating free-standing conversational groups in images,'' \emph{PloS
  one}, vol.~10, no.~5, p. e0123783, 2015.

\bibitem{zen2010space}
G.~Zen, B.~Lepri, E.~Ricci, and O.~Lanz, ``Space speaks: towards socially and
  personality aware visual surveillance,'' in \emph{Proceedings of the 1st ACM
  international workshop on Multimodal pervasive video analysis}.\hskip 1em
  plus 0.5em minus 0.4em\relax ACM, 2010, pp. 37--42.

\bibitem{vazquez2015parallel}
M.~V{\'a}zquez, A.~Steinfeld, and S.~E. Hudson, ``Parallel detection of
  conversational groups of free-standing people and tracking of their
  lower-body orientation,'' in \emph{2015 IEEE/RSJ International Conference on
  Intelligent Robots and Systems (IROS)}.\hskip 1em plus 0.5em minus
  0.4em\relax IEEE, 2015, pp. 3010--3017.

\bibitem{ciolek1983proxemics}
T.~M. Ciolek, ``The proxemics lexicon: A first approximation,'' \emph{Journal
  of Nonverbal Behavior}, vol.~8, no.~1, pp. 55--79, 1983.

\bibitem{cristani2011social}
M.~Cristani, L.~Bazzani, G.~Paggetti, A.~Fossati, D.~Tosato, A.~Del~Bue,
  G.~Menegaz, and V.~Murino, ``Social interaction discovery by statistical
  analysis of f-formations.'' in \emph{BMVC}, vol.~2, 2011, p.~4.

\bibitem{luber2013multi}
M.~Luber and K.~O. Arras, ``Multi-hypothesis social grouping and tracking for
  mobile robots.'' in \emph{Robotics: Science and Systems}, 2013.

\bibitem{setti2013multi}
F.~Setti, O.~Lanz, R.~Ferrario, V.~Murino, and M.~Cristani, ``Multi-scale
  f-formation discovery for group detection,'' in \emph{2013 IEEE International
  Conference on Image Processing}.\hskip 1em plus 0.5em minus 0.4em\relax IEEE,
  2013, pp. 3547--3551.

\bibitem{hung2011detecting}
H.~Hung and B.~Kr{\"o}se, ``Detecting f-formations as dominant sets,'' in
  \emph{Proceedings of the 13th international conference on multimodal
  interfaces}.\hskip 1em plus 0.5em minus 0.4em\relax ACM, 2011, pp. 231--238.

\bibitem{tran2013social}
K.~N. Tran, A.~Bedagkar-Gala, I.~A. Kakadiaris, and S.~K. Shah, ``Social cues
  in group formation and local interactions for collective activity analysis.''
  in \emph{VISAPP (1)}, 2013, pp. 539--548.

\bibitem{mead2013automated}
R.~Mead, A.~Atrash, and M.~J. Matari{\'c}, ``Automated proxemic feature
  extraction and behavior recognition: Applications in human-robot
  interaction,'' \emph{International Journal of Social Robotics}, vol.~5,
  no.~3, pp. 367--378, 2013.

\bibitem{gedik2018detecting}
E.~Gedik and H.~Hung, ``Detecting conversing groups using social dynamics from
  wearable acceleration: Group size awareness,'' \emph{Proceedings of the ACM
  on Interactive, Mobile, Wearable and Ubiquitous Technologies}, vol.~2, no.~4,
  p. 163, 2018.

\bibitem{swofford2020improving}
M.~Swofford, J.~Peruzzi, N.~Tsoi, S.~Thompson, R.~Mart{\'\i}n-Mart{\'\i}n,
  S.~Savarese, and M.~V{\'a}zquez, ``Improving social awareness through dante:
  Deep affinity network for clustering conversational interactants,''
  \emph{Proceedings of the ACM on Human-Computer Interaction}, vol.~4, no.
  CSCW1, pp. 1--23, 2020.

\bibitem{hedayati2019recognizing}
H.~Hedayati, D.~Szafir, and S.~Andrist, ``Recognizing f-formations in the open
  world,'' in \emph{2019 14th ACM/IEEE International Conference on Human-Robot
  Interaction (HRI)}.\hskip 1em plus 0.5em minus 0.4em\relax IEEE, 2019, pp.
  558--559.

\bibitem{alameda2015salsa}
X.~Alameda-Pineda, J.~Staiano, R.~Subramanian, L.~Batrinca, E.~Ricci, B.~Lepri,
  O.~Lanz, and N.~Sebe, ``Salsa: A novel dataset for multimodal group behavior
  analysis,'' \emph{IEEE transactions on pattern analysis and machine
  intelligence}, vol.~38, no.~8, pp. 1707--1720, 2015.

\bibitem{zen2014unsupervised}
G.~Zen, E.~Sangineto, E.~Ricci, and N.~Sebe, ``Unsupervised domain adaptation
  for personalized facial emotion recognition,'' in \emph{Proceedings of the
  16th international conference on multimodal interaction}.\hskip 1em plus
  0.5em minus 0.4em\relax ACM, 2014, pp. 128--135.

\bibitem{bohus2014}
D.~Bohus, C.~W. Saw, and E.~Horvitz, ``Directions robot: in-the-wild
  experiences and lessons learned,'' in \emph{In Proceedings of the 2014
  international conference on Autonomous agents and multi-agent systems}.\hskip
  1em plus 0.5em minus 0.4em\relax IFAAMAS, 2014, pp. 637--644.

\end{thebibliography}

\end{document}